\documentclass[a4paper]{article}

\usepackage{INTERSPEECH2018,empheq,url,booktabs,multirow,cite}

\title{Multi-Head Decoder for End-to-End Speech Recognition}
\name{Tomoki Hayashi$^1$, Shinji Watanabe$^2$, Tomoki Toda$^1$, Kazuya Takeda$^1$}
\address{$^1$Nagoya University, $^2$Johns Hopkins University}
\email{hayashi.tomoki@g.sp.m.is.nagoya-u.ac.jp, shinjiw@jhu.edu,\\
       tomoki@icts.nagoya-u.ac.jp, kazuya.takeda@nagoya-u.ac.jp}

\begin{document}

\maketitle
\begin{abstract}
This paper presents a new network architecture called multi-head decoder for end-to-end speech recognition as an extension of a multi-head attention model.
In the multi-head attention model, multiple attentions are calculated, and then, they are integrated into a single attention.
On the other hand, instead of the integration in the attention level, our proposed method uses multiple decoders for each attention and integrates their outputs to generate a final output.
Furthermore, in order to make each head to capture the different modalities, different attention functions are used for each head, leading to the improvement of the recognition performance with an ensemble effect.
To evaluate the effectiveness of our proposed method, we conduct an experimental evaluation using Corpus of Spontaneous Japanese.
Experimental results demonstrate that our proposed method outperforms the conventional methods such as location-based and multi-head attention models, and that it can capture different speech/linguistic contexts within the attention-based encoder-decoder framework.
\end{abstract}


\section{Introduction}
Automatic speech recognition (ASR) is the task to convert a continuous speech signal into a sequence of discrete characters, and it is a key technology to realize the interaction between human and machine.
ASR has a great potential for various applications such as voice search and voice input, making our lives more rich.
Typical ASR systems~\cite{jelinek1976continuous} consist of many modules such as an acoustic model, a lexicon model, and a language model.
Factorizing the ASR system into these modules makes it possible to deal with each module as a separate problem.
Over the past decades, this factorization has been the basis of the ASR system, however, it makes the system much more complex.

With the improvement of deep learning techniques, end-to-end approaches have been proposed~\cite{chorowski2014end}.
In the end-to-end approach, a continuous acoustic signal or a sequence of acoustic features is directly converted into a sequence of characters with a single neural network.
Therefore, the end-to-end approach does not require the factorization into several modules, as described above, making it easy to optimize the whole system.
Furthermore, it does not require lexicon information, which is handcrafted by human experts in general.

The end-to-end approach is classified into two types.
One approach is based on connectionist temporal classification (CTC)~\cite{graves2006connectionist,graves2014towards,chorowski2014end}, which makes it possible to handle the difference in the length of input and output sequences with dynamic programming.
The CTC-based approach can efficiently solve the sequential problem, however, CTC uses Markov assumptions to perform dynamic programming and predicts output symbols such as characters or phonemes for each frame independently.
Consequently, except in the case of huge training data~\cite{amodei2016deep,soltau2016neural}, it requires the language model and graph-based decoding~\cite{miao2015eesen}.

The other approach utilizes attention-based method~\cite{chorowski2015attention}.
In this approach, encoder-decoder architecture~\cite{cho2014learning,sutskever2014sequence} is used to perform a direct mapping from a sequence of input features into text.
The encoder network converts the sequence of input features to that of discriminative hidden states, and the decoder network uses attention mechanism to get an alignment between each element of the output sequence and the encoder hidden states.
And then it estimates the output symbol using weighted averaged hidden states, which is based on the alignment, as the inputs of the decoder network.
Compared with the CTC-based approach, the attention-based method does not require any conditional independence assumptions including the Markov assumption, language models, and complex decoding.
However, non-causal alignment problem is caused by a too flexible alignment of the attention mechanism~\cite{watanabe2017hybrid}.
To address this issue, the study~\cite{watanabe2017hybrid} combines the objective function of the attention-based model with that of CTC to constrain flexible alignments of the attention.
Another study~\cite{chiu2017state} uses a multi-head attention (MHA) to get more suitable alignments.
In MHA, multiple attentions are calculated, and then, they are integrated into a single attention. 
Using MHA enables the model to jointly focus on information from different representation subspaces at different positions~\cite{vaswani2017attention}, leading to the improvement of the recognition performance.
 
Inspired by the idea of MHA, in this study we present a new network architecture called multi-head decoder for end-to-end speech recognition as an extension of a multi-head attention model.
Instead of the integration in the attention level, our proposed method uses multiple decoders for each attention and integrates their outputs to generate a final output.
Furthermore, in order to make each head to capture the different modalities, different attention functions are used for each head, leading to the improvement of the recognition performance with an ensemble effect.
To evaluate the effectiveness of our proposed method, we conduct an experimental evaluation using Corpus of Spontaneous Japanese.
Experimental results demonstrate that our proposed method outperforms the conventional methods such as location-based and multi-head attention models, and that it can capture different speech/linguistic contexts within the attention-based encoder-decoder framework.

\section{Attention-Based End-to-End ASR}
\begin{figure}[t]
\begin{center}
\includegraphics[width=0.9\columnwidth]{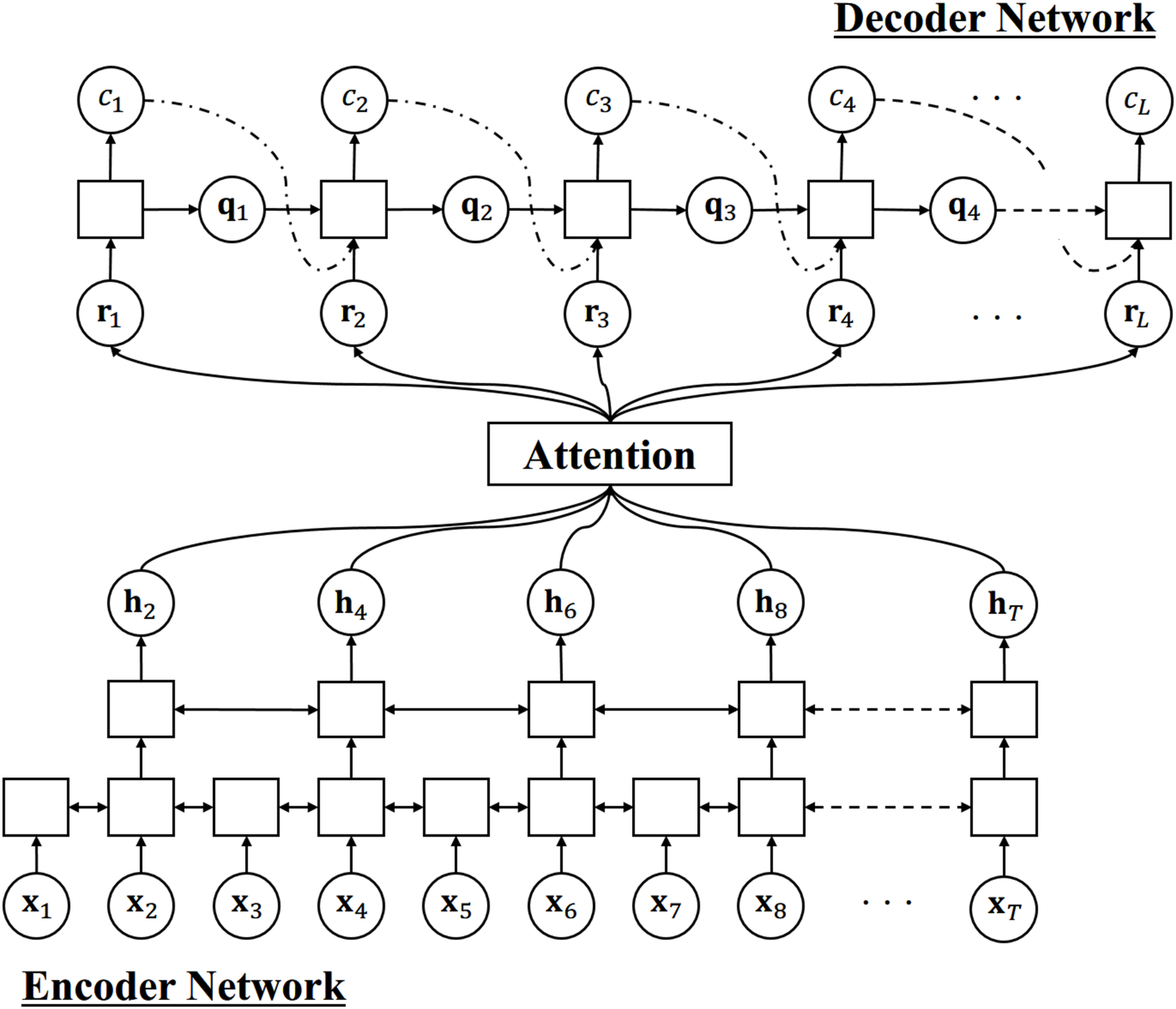}
\end{center}
\vspace{-2mm}
\caption{\it Overview of attention-based network architecture.}
\label{fig:att_overview}
\end{figure}

The overview of attention-based network architecture is shown in Fig.~\ref{fig:att_overview}.
The attention-based method directly estimates a posterior $p(\mathbf{C}|\mathbf{X})$, where $\mathbf{X} = \{\mathbf{x}_1,\mathbf{x}_2, \dots, \mathbf{x}_T\}$ represents a sequence of input features, $\mathbf{C} = \{c_1, c_2, \dots, c_L\}$ represents a sequence of output characters. 
The posterior $p(\mathbf{C}|\mathbf{X})$ is factorized with a probabilistic chain rule as follows:
\begin{equation}
    \label{eq:att_posterior}
    p(\mathbf{C}|\mathbf{X}) = \displaystyle\prod_{l=1}^L p(c_l | c_{1:l-1}, \mathbf{X}),
\end{equation}
where $c_{1:l-1}$ represents a subsequence $\{c_1, c_2, \dots\, c_{l-1}\}$, and $p(c_l | c_{1:l-1}, \mathbf{X})$ is calculated as follows:
\begin{eqnarray}
    \label{eq:att_encoder}
    & \mathbf{h}_t = \mathrm{Encoder}(\mathbf{X}),
    \vspace{-3mm}
\end{eqnarray}
\begin{eqnarray}
    \label{eq:att_weight1}
    & a_{lt} = 
    \begin{cases}
        &\mathrm{DotProductAttention}(\mathbf{q}_{l-1}, \mathbf{h}_t), \\
        &\mathrm{AdditiveAttention}(\mathbf{q}_{l-1}, \mathbf{h}_t), \\
        &\mathrm{LocationAttention}(\mathbf{q}_{l-1}, \mathbf{h}_t, \mathbf{a}_{l-1}), \\
        &\mathrm{CoverageAttention}(\mathbf{q}_{l-1}, \mathbf{h}_t, \mathbf{a}_{1:l-1}), \\
    \end{cases}
\vspace{-3mm}
\end{eqnarray}
\begin{eqnarray}
    \label{eq:att_weight2}
    & \mathbf{r}_l = \displaystyle\sum_{t=1}^{T} a_{lt}\mathbf{h}_t, \\
    \label{eq:att_decoder}
    & p(c_l | c_{1:l-1}, \mathbf{X}) = \mathrm{Decoder}(\mathbf{r}_l, \mathbf{q}_{l-1}, c_{l-1}),
\end{eqnarray}
where Eq.~(\ref{eq:att_encoder}) and Eq.~(\ref{eq:att_decoder}) represent encoder and decoder networks, respectively, $a_{lt}$ represents an attention weight, $\mathbf{a}_l$ represents an attention weight vector, which is a sequence of attention weights $\{a_{l0}, a_{l1}, \dots, a_{lT}\}$, $\mathbf{a}_{1:l-1}$ represents a subsequence of attention vectors $\{\mathbf{a}_{1}, \mathbf{a}_{2}, \dots, \mathbf{a}_{l-1}\}$, $\mathbf{h}_t$ and $\mathbf{q}_l$ represent hidden states of encoder and decoder networks, respectively, and $\mathbf{r}_l$ represents the letter-wise hidden vector, which is a weighted summarization of hidden vectors with the attention weight vector $\mathbf{a}_l$.

The encoder network in Eq.~(\ref{eq:att_encoder}) converts a sequence of input features $\mathbf{X}$ into frame-wise discriminative hidden states $\mathbf{h}_t$, and it is typically modeled by a bidirectional long short-term memory recurrent neural network (BLSTM):
\begin{equation}
    \label{eq:att_encoder_blstm}
    \mathrm{Encoder}(\mathbf{X}) = \mathrm{BLSTM}(\mathbf{X}).
\end{equation}
In the case of ASR, the length of the input sequence is significantly different from the length of the output sequence.
Hence, basically outputs of BLSTM are often subsampled to reduce the computational cost~\cite{chorowski2015attention,chan2016listen}. 

The attention weight $a_{lt}$ in Eq.~(\ref{eq:att_weight1}) represents a soft alignment between each element of the output sequence $c_l$ and the encoder hidden states $\mathbf{h}_t$.
\begin{itemize}
\setlength{\leftskip}{-5mm}
\item $\mathrm{DotProductAttention}(\cdot)$ in Eq.~(\ref{eq:att_weight1}), which is the most simplest attention~\cite{luong2015effective}, is calculated as follows:
\begin{eqnarray}
    \label{eq:dot_att}
    & e_{lt} = \mathbf{q}_{l-1}^\mathrm{T}\mathbf{W}_a\mathbf{h}_t, \\
    & \mathbf{a}_{l} = \mathrm{Softmax}(\mathbf{e}_l),
\end{eqnarray}
where $\mathbf{W}_a$ represents trainable matrix parameters, and $\mathbf{e}_l$ represent a sequence of energies $\{e_{l1}, e_{l2}, \dots, e_{lT}\}$
\item $\mathrm{AdditiveAttention}(\cdot)$ in Eq.~(\ref{eq:att_weight1}) is additive attention~\cite{bahdanau2014neural}, and the calculation of the energy in Eq.~(\ref{eq:dot_att}) is replaced with following equation:%
\begin{equation}
    \label{eq:add_att}
    e_{lt} = \mathbf{g}^\mathrm{T}\tanh(\mathbf{W}_q\mathbf{q}_{l-1} + \mathbf{W}_h\mathbf{h}_t + \mathbf{b}),
\end{equation}
where $\mathbf{W}_*$ represents trainable matrix parameters, $\mathbf{g}$ and $\mathbf{b}$ represent trainable vector parameters.
\item $\mathrm{LocationAttention}(\cdot)$ in Eq.~(\ref{eq:att_weight1}) is location-based attention~\cite{chorowski2015attention}, and the calculation of the energy in Eq.~(\ref{eq:dot_att}) is replaced with following equations:
\begin{equation}
    \label{eq:add_loc}
    \mathbf{F}_l = \mathbf{K} * \mathbf{a}_{l-1},
\end{equation}
\vspace{-5mm}
\begin{equation}
    e_{lt} = \mathbf{g}^\mathrm{T}\tanh(\mathbf{W}_q\mathbf{q}_{l-1} + \mathbf{W}_h\mathbf{h}_t + \mathbf{W}_f\mathbf{f}_{lt} + \mathbf{b}),
\vspace{2mm}
\end{equation}
where $\mathbf{F}_l$ consists of the vectors $\{\mathbf{f}_{l1}, \mathbf{f\hspace{1mm}}_{l2}, \dots, \mathbf{f}_{lT} \}$, and $\mathbf{K}$ represents trainable one-dimensional convolution filters.
\item $\mathrm{CoverageAttention}(\cdot)$ in Eq.~(\ref{eq:att_weight1}) is coverage mechanism~\cite{see2017get}, and the calculation of the energy in Eq.~(\ref{eq:dot_att}) is replaced with following equations:
\begin{equation}
    \label{eq:add_cov}
    \mathbf{v}_l = \displaystyle\sum_{l'=1}^{l-1} \mathbf{a}_{l'},
\end{equation}
\vspace{-3mm}
\begin{equation}
    e_{lt} = \mathbf{g}^\mathrm{T}\tanh(\mathbf{W}_q\mathbf{q}_{l-1} + \mathbf{W}_h\mathbf{h}_t + \mathbf{w}_v v_{lt} + \mathbf{b}),
\vspace{2mm}
\end{equation}
where $\mathbf{w}$ represents trainable vector parameters.
\end{itemize}

The decoder network in Eq.~(\ref{eq:att_decoder}) estimates the next character $c_l$ from the previous character $c_{l-1}$, hidden state vector of itself $\mathbf{q}_{l-1}$ and the letter-wise hidden state vector $\mathbf{r}_l$, similar to RNN language model (RNNLM)~\cite{mikolov2010recurrent}.
It is typically modeled using LSTM as follows:
\begin{eqnarray}
    \label{eq:att_decoder_lstm}
    & \mathbf{q}_l = \mathrm{LSTM}(c_{l-1}, \mathbf{q}_{l-1}, \mathbf{r}_l), \\
    & \mathrm{Decoder}(\cdot) = \mathrm{Softmax}(\mathbf{W}\mathbf{q}_l+\mathbf{b}),
\end{eqnarray}
where $\mathbf{W}$ and $\mathbf{b}$ represent trainable matrix and vector parameters, respectively.

Finally, the whole of above networks are optimized using back-propagation through time (BPTT)~\cite{werbos1990backpropagation} to minimize the following objective function:
\begin{equation}
\begin{split}
    \label{eq:att_obj}
    \mathcal{L} & = - \log p(\mathbf{C} | \mathbf{X}) \\
                & = - \log\left(\sum_{l=1}^L p(c_l | c_{1:l-1}^*, \mathbf{X})\right),
\end{split}
\end{equation}
where $c_{1:l-1}^* = \{c_1^*, c_2^*, \dots, c_{l-1}^*\}$ represents the ground truth of the previous characters.

\section{Multi-Head Decoder}
\begin{figure}[t]
\begin{center}
\includegraphics[width=1.05\columnwidth]{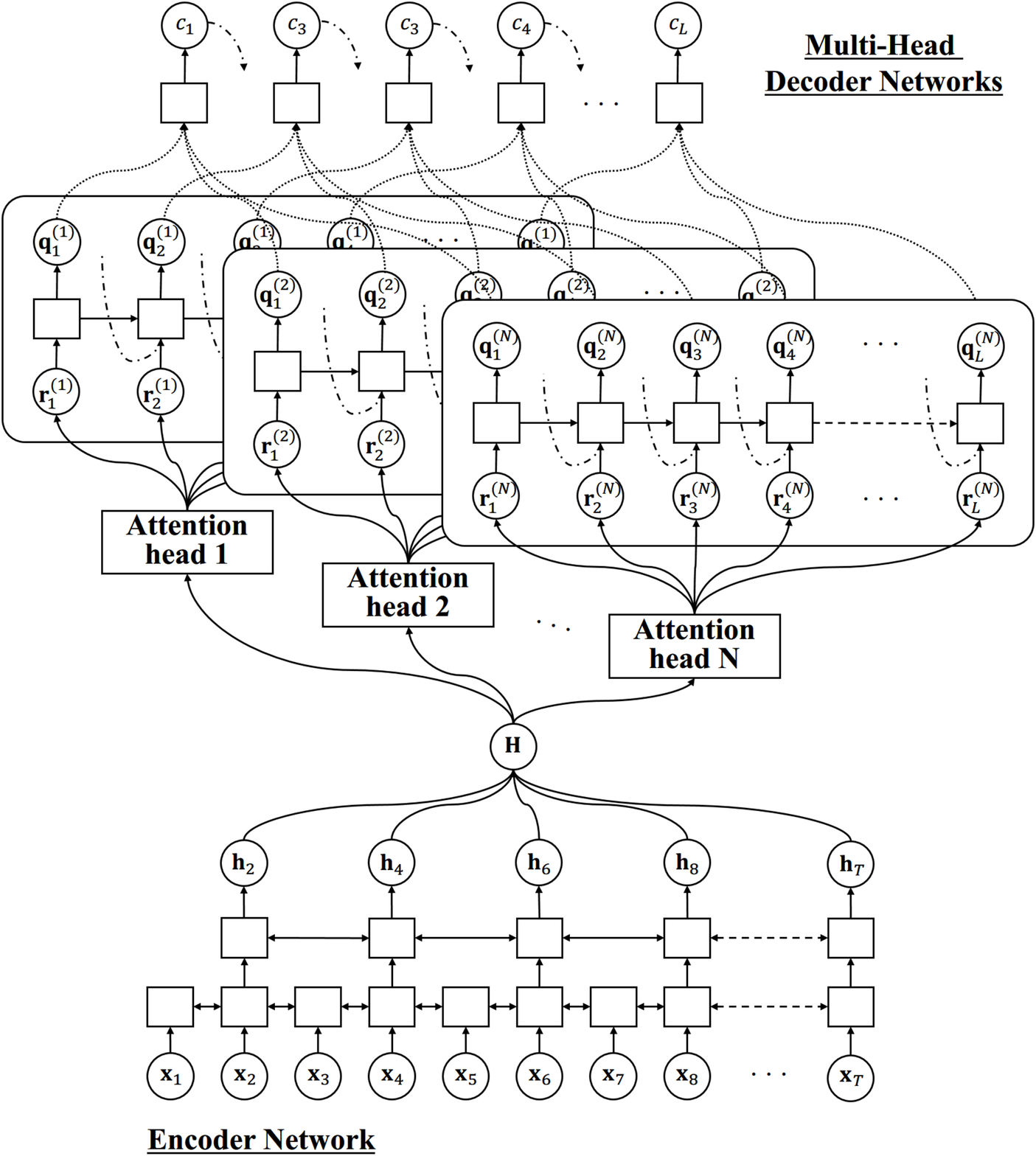}
\end{center}
\vspace{-2mm}
\caption{\it Overview of multi-head decoder architecture.}
\label{fig:overview}
\end{figure}
The overview of our proposed multi-head decoder (MHD) architecture is shown in Fig.~\ref{fig:overview}.
In MHD architecture, multiple attentions are calculated with the same manner in the conventional multi-head attention (MHA)~\cite{vaswani2017attention}.
We first describe the conventional MHA, and extend it to our proposed multi-head decoder (MHD).

\subsection{Multi-head attention (MHA)}
The layer-wise hidden vector at the head $n$ is calculated as follows:
\begin{eqnarray}
    \label{eq:mha}
    & a_{lt}^{(n)} = \mathrm{Attention}(\mathbf{W}_Q^{(n)}\mathbf{q}_{l-1}, \mathbf{W}_K^{(n)}\mathbf{h}_t, \dots), \\
    & \mathbf{r}_l^{(n)} = \displaystyle\sum_{t=1}^{T} a_{lt}^{(n)}\mathbf{W}_V^{(n)}\mathbf{h}_t,
\end{eqnarray}
where $\mathbf{W}_Q^{(n)}$, $\mathbf{W}_K^{(n)}$, and $\mathbf{W}_V^{(n)}$ represent trainable matrix parameters, and any types of attention in Eq.~(\ref{eq:att_weight1}) can be used for $\mathrm{Attention}(\cdot)$ in Eq.~(\ref{eq:mha}).

In the case of MHA, the layer-wise hidden vectors of each head are integrated into a single vector with a trainable linear transformation:
\begin{equation}
    \mathbf{r}_l = \mathbf{W}_O\left[{\mathbf{r}^{(1)}_l}^\top, {\mathbf{r}^{(2)}_l}^\top, \dots, {\mathbf{r}^{(N)}_l}^\top\right]^\top,
\end{equation}
where $\mathbf{W}_O$ is a trainable matrix parameter, $N$ represents the number of heads.

\subsection{Multi-head decoder (MHD)}
On the other hand, in the case of MHD, instead of the integration at attention level, we assign multiple decoders for each head and then integrate their outputs to get a final output.
Since each attention decoder captures different modalities, it is expected to improve the recognition performance with an ensemble effect.
The calculation of the attention weight at the head $n$ in Eq.~(\ref{eq:mha}) is replaced with following equation:%
\begin{equation}
    a_{lt}^{(n)} = \mathrm{Attention}(\mathbf{W}_Q^{(n)}\mathbf{q}^{(n)}_{l-1}, \mathbf{W}_K^{(n)}\mathbf{h}_t, \dots).
\end{equation}
Instead of the integration of the letter-wise hidden vectors $\{\mathbf{r}_l^{(1)}, \mathbf{r}_l^{(2)}, \dots, \mathbf{r}_l^{(N)}\}$ with linear transformation, each letter-wise hidden vector $\mathbf{r}_l^{(n)}$ is fed to $n$-th decoder LSTM:%
\begin{equation}
    \mathbf{q}_l^{(n)} = \mathrm{LSTM}^{(n)}(c_{l-1}, \mathbf{q}_{l-1}^{(n)}, \mathbf{r}_l^{(n)}).
\end{equation}
Note that each LSTM has its own hidden state $\mathbf{q}_l^{(n)}$ which is used for the calculation of the attention weight $a_{lt}^{(n)}$, while the input character $c_l$ is the same among all of the LSTMs.
Finally, all of the outputs are integrated as follows:
\begin{eqnarray}
    & p(c_l | c_{1:l-1}, \mathbf{X}) = \mathrm{Softmax}\left(\displaystyle\sum_{n=1}^{N}\mathbf{W}^{(n)}\mathbf{q}_{l}^{(n)}+\mathbf{b}\right),
\end{eqnarray}
where $\mathbf{W}^{(n)}$ represents a trainable matrix parameter, and $\mathbf{b}$ represents a trainable vector parameter.

\subsection{Heterogeneous multi-head decoder (HMHD)}
\label{sec:hmhd}
As a further extension, we propose heterogeneous multi-head decoder (HMHD).
Original MHA methods~\cite{vaswani2017attention,chiu2017state} use the same attention function such as dot-product or additive attention for each head.
On the other hand, HMHD uses different attention functions for each head.
We expect that this extension enables to capture the further different context in speech within the attention-based encoder-decoder framework.

\section{Experimental Evaluation}
\begin{table}[t]
\begin{center}
\caption{Experimental conditions.}
\vspace{0mm}
\label{tb:cond}
\scalebox{1}{%
{\renewcommand\arraystretch{1.0}
\begin{tabular}{l l} \toprule
\ \# training                    & \ 445,068 utterances (581 hours) \\
\ \# evaluation (task 1)         & \ 1,288 utterances (1.9 hours) \\
\ \# evaluation (task 2)         & \ 1,305 utterances (2.0 hours) \\
\ \# evaluation (task 3)         & \ 1,389 utterances (1.3 hours) \\ \midrule \midrule
\ Sampling rate                  & \ 16,000 Hz \\
\ Window size                    & \ 25 ms \\
\ Shift size                     & \ 10 ms \\ \midrule \midrule
\ Encoder type                   & \ BLSTMP \\
\ \# encoder layers              & \ 6 \\
\ \# encoder units               & \ 320 \\
\ \# projection units            & \ 320 \\
\ Decoder type                   & \ LSTM \\
\ \# decoder layers              & \ 1 \\
\ \# decoder units               & \ 320 \\
\ \# heads in MHA                & \ 4 \\
\ \# filter in location att.     & \ 10 \\
\ Filter size in location att.   & \ 100 \\ \midrule \midrule
\ Learning rate                  & \ 1.0 \\
\ Initialization                 & \ Uniform $[-0.1, 0.1]$ \\
\ Gradient clipping norm         & \ 5 \\
\ Batch size                     & \ 30 \\
\ Maximum epoch                  & \ 15 \\
\ Optimization method            & \ AdaDelta~\cite{zeiler2012adadelta} \\
\ AdaDelta $\rho$                & \ 0.95 \\
\ AdaDelta $\epsilon$            & \ $10^{-8}$ \\
\ AdaDelta $\epsilon$ decay rate & \ $10^{-2}$ \\ \midrule \midrule
\ Beam size                      & \ 20 \\
\ Maximum length                 & \ 0.5 \\
\ Minimum length                 & \ 0.1 \\ \bottomrule
\end{tabular}
}}
\vspace{0mm}
\end{center}
\end{table}
To evaluate the performance of our proposed method, we conducted experimental evaluation using Corpus of Spontaneous Japanese (CSJ)~\cite{maekawa2000spontaneous}, including 581 hours of training data, and three types of evaluation data.
To compare the performance, we used following dot, additive, location, and three variants of multi-head attention methods:
\begin{itemize}
    \setlength{\parskip}{0mm} 
    \setlength{\itemsep}{-2mm} 
    \item Dot-product attention-based model (Dot),\\
    \item Additive attention-based model (Add),\\
    \item Location-aware attention-based model (Loc),\\
    \item Multi-head dot-product attention model (MHA-Dot),\\
    \item Multi-head additive attention model (MHA-Add),\\
    \item Multi-head location attention model (MHA-Loc).
\end{itemize}
We used the input feature vector consisting of 80 dimensional log Mel filter bank and three dimensional pitch feature, which is extracted using open-source speech recognition toolkit Kaldi~\cite{povey2011kaldi}.
Encoder and decoder networks were six-layered BLSTM with projection layer~\cite{sak2014long} (BLSTMP) and one-layered LSTM, respectively.
In the second and third bottom layers in the encoder, subsampling was performed to reduce the length of utterance, yielding the length $T/4$.
For MHA/MHD, we set the number of heads to four.
For HMHD, we used two kind of settings: (1) dot-product attention + additive attention + location-based attention + coverage mechanism attention (Dot+Add+Loc+Cov), and (2) two location-based attentions + two coverage mechanism attentions (2$\times$Loc+2$\times$Cov).
The number of distinct output characters was 3,315 including Kanji, Hiragana, Katakana, alphabets, Arabic number and sos/eos symbols.
In decoding, we used beam search algorithm~\cite{sutskever2014sequence} with beam size 20.
We manually set maximum and minimum lengths of the output sequence to 0.1 and 0.5 times the length of the subsampled input sequence, respectively, and the length penalty to 0.1 times the length of the output sequence.
All of the networks were trained using end-to-end speech processing toolkit ESPnet~\cite{espnet} with a single GPU (Titan X pascal).
Character error rate (CER) was used as a metric.
The detail of experimental condition is shown in Table~\ref{tb:cond}.

Experimental results are shown in Table~\ref{tb:results}.
\begin{table}[t]
\begin{center}
\vspace{0mm}
\caption{Experimental results.}
\label{tb:results}
\vspace{-0mm}
\scalebox{1}{%
{\renewcommand\arraystretch{1.0}
\begin{tabular}{l c c c}
                                & \multicolumn{3}{c}{CER [\%]} \\ \cmidrule{2-4}
                                & Task 1     & Task 2    & Task 3    \\ \toprule
Dot                             & 12.7       & 9.8       & 10.7      \\
Add                             & 11.1       & 8.4       & 9.0      \\
Loc                             & 11.7       & 8.8       & 10.2      \\
MHA-Dot                         & 11.6       & 8.5       & 9.3       \\
MHA-Add                         & 10.7       & 8.2       & 9.1       \\
MHA-Loc                         & 11.5       & 8.6       & 9.0       \\ \midrule \midrule
MHD-Loc                         & 11.0       & 8.4       & 9.5       \\
HMHD (Dot+Add+Loc+Cov)           & 11.0       & 8.3       & 9.0       \\
HMHD (2$\times$Loc+2$\times$Cov) & {\bf 10.4} & {\bf 7.7} & {\bf 8.9} \\ \bottomrule
\end{tabular}
}}
\vspace{0mm}
\end{center}
\end{table}
First, we focus on the results of the conventional methods.
Basically, it is known that location-based attention yields better performance than additive attention~\cite{watanabe2017hybrid}.
However, in the case of Japanese sentence, its length is much shorter than that of English sentence, which makes the use of location-based attention less effective.
In most of the cases, the use of MHA brings the improvement of the recognition performance.
Next, we focus on the effectiveness of our proposed MHD architecture.
By comparing with the MHA-Loc, MHD-Loc (proposed method) improved the performance in Tasks 1 and 2, while we observed the degradation in Task 3.
However, the heterogeneous extension (HMHD), as introduced in Section~\ref{sec:hmhd}, brings the further improvement for the performance of MHD, achieving the best performance among all of the methods for all test sets.

\begin{figure}[t]
\begin{center}
\includegraphics[width=1.0\columnwidth]{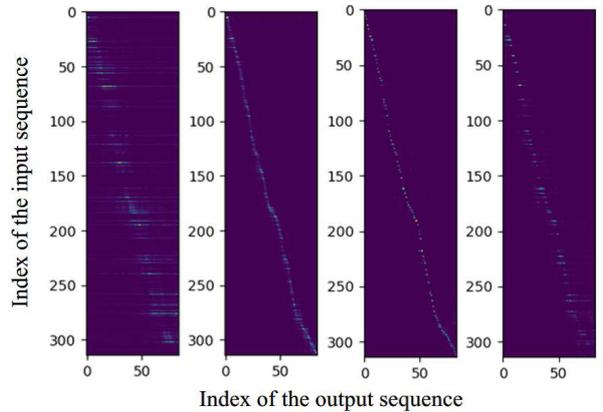}
\end{center}
\vspace{-3mm}
\caption{\it Attention weights of each head. Two left figures represent the attention weights of the location-based attention, and the remaining figures represent that of the coverage mechanism attention.}
\label{fig:att_weights}
\end{figure}

Finally, Figure~\ref{fig:att_weights} shows the alignment information of each head of HMHD (2$\times$Loc+2$\times$Cov), which was obtained by visualizing the attention weights.
Interestingly, the alignments of the right and left ends seem to capture more abstracted dynamics of speech, while the rest of two alignments behave like normal alignments obtained by a standard attention mechanism.
Thus, we can see that the attention weights of each head have a different tendency, and it supports our hypothesis that HMHD can capture different speech/linguistic contexts within its framework.

\section{Conclusions}
In this paper, we proposed a new network architecture called multi-head decoder for end-to-end speech recognition as an extension of a multi-head attention model.
Instead of the integration in the attention level, our proposed method utilized multiple decoders for each attention and integrated their outputs to generate a final output.
Furthermore, in order to make each head to capture the different modalities, we used different attention functions for each head.
To evaluate the effectiveness of our proposed method, we conducted an experimental evaluation using Corpus of Spontaneous Japanese.
Experimental results demonstrated that our proposed methods outperformed the conventional methods such as location-based and multi-head attention models, and that it could capture different speech/linguistic contexts within the attention-based encoder-decoder framework.

In the future work, we will combine the multi-head decoder architecture with Joint CTC/Attention architecture~\cite{watanabe2017hybrid}, and evaluate the performance using other databases.

\bibliographystyle{IEEEtran}
\bibliography{mybib}

\begin{thebibliography}{10}
\providecommand{\url}[1]{#1}
\csname url@samestyle\endcsname
\providecommand{\newblock}{\relax}
\providecommand{\bibinfo}[2]{#2}
\providecommand{\BIBentrySTDinterwordspacing}{\spaceskip=0pt\relax}
\providecommand{\BIBentryALTinterwordstretchfactor}{4}
\providecommand{\BIBentryALTinterwordspacing}{\spaceskip=\fontdimen2\font plus
\BIBentryALTinterwordstretchfactor\fontdimen3\font minus
  \fontdimen4\font\relax}
\providecommand{\BIBforeignlanguage}[2]{{%
\expandafter\ifx\csname l@#1\endcsname\relax
\typeout{** WARNING: IEEEtran.bst: No hyphenation pattern has been}%
\typeout{** loaded for the language `#1'. Using the pattern for}%
\typeout{** the default language instead.}%
\else
\language=\csname l@#1\endcsname
\fi
#2}}
\providecommand{\BIBdecl}{\relax}
\BIBdecl

\bibitem{jelinek1976continuous}
F.~Jelinek, ``Continuous speech recognition by statistical methods,''
  \emph{Proceedings of the IEEE}, vol.~64, no.~4, pp. 532--556, 1976.

\bibitem{chorowski2014end}
J.~Chorowski, D.~Bahdanau, K.~Cho, and Y.~Bengio, ``End-to-end continuous
  speech recognition using attention-based recurrent nn: First results,''
  \emph{arXiv preprint arXiv:1412.1602}, 2014.

\bibitem{graves2006connectionist}
A.~Graves, S.~Fern{\'a}ndez, F.~Gomez, and J.~Schmidhuber, ``Connectionist
  temporal classification: labelling unsegmented sequence data with recurrent
  neural networks,'' in \emph{Proceedings of the 23rd international conference
  on Machine learning}.\hskip 1em plus 0.5em minus 0.4em\relax ACM, 2006, pp.
  369--376.

\bibitem{graves2014towards}
A.~Graves and N.~Jaitly, ``Towards end-to-end speech recognition with recurrent
  neural networks,'' in \emph{International Conference on Machine Learning},
  2014, pp. 1764--1772.

\bibitem{amodei2016deep}
D.~Amodei, S.~Ananthanarayanan, R.~Anubhai, J.~Bai, E.~Battenberg, C.~Case,
  J.~Casper, B.~Catanzaro, Q.~Cheng, G.~Chen \emph{et~al.}, ``Deep speech 2:
  End-to-end speech recognition in english and mandarin,'' in
  \emph{International Conference on Machine Learning}, 2016, pp. 173--182.

\bibitem{soltau2016neural}
H.~Soltau, H.~Liao, and H.~Sak, ``Neural speech recognizer: Acoustic-to-word
  lstm model for large vocabulary speech recognition,'' \emph{arXiv preprint
  arXiv:1610.09975}, 2016.

\bibitem{miao2015eesen}
Y.~Miao, M.~Gowayyed, and F.~Metze, ``Eesen: End-to-end speech recognition
  using deep rnn models and wfst-based decoding,'' in \emph{IEEE Workshop on
  Automatic Speech Recognition and Understanding (ASRU)}.\hskip 1em plus 0.5em
  minus 0.4em\relax IEEE, 2015, pp. 167--174.

\bibitem{chorowski2015attention}
J.~K. Chorowski, D.~Bahdanau, D.~Serdyuk, K.~Cho, and Y.~Bengio,
  ``Attention-based models for speech recognition,'' in \emph{Advances in
  neural information processing systems}, 2015, pp. 577--585.

\bibitem{cho2014learning}
K.~Cho, B.~Van~Merri{\"e}nboer, C.~Gulcehre, D.~Bahdanau, F.~Bougares,
  H.~Schwenk, and Y.~Bengio, ``Learning phrase representations using rnn
  encoder-decoder for statistical machine translation,'' \emph{arXiv preprint
  arXiv:1406.1078}, 2014.

\bibitem{sutskever2014sequence}
I.~Sutskever, O.~Vinyals, and Q.~V. Le, ``Sequence to sequence learning with
  neural networks,'' in \emph{Advances in neural information processing
  systems}, 2014, pp. 3104--3112.

\bibitem{watanabe2017hybrid}
S.~Watanabe, T.~Hori, S.~Kim, J.~R. Hershey, and T.~Hayashi, ``Hybrid
  ctc/attention architecture for end-to-end speech recognition,'' \emph{IEEE
  Journal of Selected Topics in Signal Processing}, vol.~11, no.~8, pp.
  1240--1253, 2017.

\bibitem{chiu2017state}
C.-C. Chiu, T.~N. Sainath, Y.~Wu, R.~Prabhavalkar, P.~Nguyen, Z.~Chen,
  A.~Kannan, R.~J. Weiss, K.~Rao, K.~Gonina \emph{et~al.}, ``State-of-the-art
  speech recognition with sequence-to-sequence models,'' \emph{arXiv preprint
  arXiv:1712.01769}, 2017.

\bibitem{vaswani2017attention}
A.~Vaswani, N.~Shazeer, N.~Parmar, J.~Uszkoreit, L.~Jones, A.~N. Gomez,
  {\L}.~Kaiser, and I.~Polosukhin, ``Attention is all you need,'' in
  \emph{Advances in Neural Information Processing Systems}, 2017, pp.
  6000--6010.

\bibitem{chan2016listen}
W.~Chan, N.~Jaitly, Q.~Le, and O.~Vinyals, ``Listen, attend and spell: A neural
  network for large vocabulary conversational speech recognition,'' in
  \emph{IEEE International Conference on Acoustics, Speech and Signal
  Processing (ICASSP)}.\hskip 1em plus 0.5em minus 0.4em\relax IEEE, 2016, pp.
  4960--4964.

\bibitem{luong2015effective}
M.-T. Luong, H.~Pham, and C.~D. Manning, ``Effective approaches to
  attention-based neural machine translation,'' \emph{arXiv preprint
  arXiv:1508.04025}, 2015.

\bibitem{bahdanau2014neural}
D.~Bahdanau, K.~Cho, and Y.~Bengio, ``Neural machine translation by jointly
  learning to align and translate,'' \emph{arXiv preprint arXiv:1409.0473},
  2014.

\bibitem{see2017get}
A.~See, P.~J. Liu, and C.~D. Manning, ``Get to the point: Summarization with
  pointer-generator networks,'' \emph{arXiv preprint arXiv:1704.04368}, 2017.

\bibitem{mikolov2010recurrent}
T.~Mikolov, M.~Karafi{\'a}t, L.~Burget, J.~{\v{C}}ernock{\`y}, and
  S.~Khudanpur, ``Recurrent neural network based language model,'' in
  \emph{Eleventh Annual Conference of the International Speech Communication
  Association}, 2010.

\bibitem{werbos1990backpropagation}
P.~J. Werbos, ``Backpropagation through time: what it does and how to do it,''
  \emph{Proceedings of the IEEE}, vol.~78, no.~10, pp. 1550--1560, 1990.

\bibitem{zeiler2012adadelta}
M.~D. Zeiler, ``Adadelta: an adaptive learning rate method,'' \emph{arXiv
  preprint arXiv:1212.5701}, 2012.

\bibitem{maekawa2000spontaneous}
K.~Maekawa, H.~Koiso, S.~Furui, and H.~Isahara, ``Spontaneous speech corpus of
  japanese.'' in \emph{LREC}.\hskip 1em plus 0.5em minus 0.4em\relax Citeseer,
  2000.

\bibitem{povey2011kaldi}
D.~Povey, A.~Ghoshal, G.~Boulianne, L.~Burget, O.~Glembek, N.~Goel,
  M.~Hannemann, P.~Motlicek, Y.~Qian, P.~Schwarz \emph{et~al.}, ``The kaldi
  speech recognition toolkit,'' in \emph{IEEE 2011 workshop on automatic speech
  recognition and understanding}, no. EPFL-CONF-192584.\hskip 1em plus 0.5em
  minus 0.4em\relax IEEE Signal Processing Society, 2011.

\bibitem{sak2014long}
H.~Sak, A.~Senior, and F.~Beaufays, ``Long short-term memory recurrent neural
  network architectures for large scale acoustic modeling,'' in \emph{Fifteenth
  annual conference of the international speech communication association},
  2014.

\bibitem{espnet}
``{ESPnet},'' \url{https://github.com/espnet/espnet}.

\end{thebibliography}
\end{document}